\documentclass[10pt,twocolumn]{article}

\usepackage[utf8]{inputenc}
\usepackage[T1]{fontenc}
\usepackage{times}

\usepackage{geometry}
\usepackage{graphicx}
\usepackage{tikz}
\usetikzlibrary{arrows.meta, positioning, shapes, fit, backgrounds, calc}
\usepackage{pgfplots}
\pgfplotsset{compat=1.16}

\usepackage{amsmath,amsfonts,amssymb}

\usepackage{booktabs}
\usepackage{multirow}
\usepackage{tabularx}
\usepackage{array}

\usepackage[authoryear,sort]{natbib}
\usepackage{hyperref}
\hypersetup{
  colorlinks=true,
  linkcolor=blue,
  citecolor=blue,
  urlcolor=blue,
  pdfborder={0 0 0}
}

\usepackage{microtype}
\usepackage{balance}
\usepackage{dblfloatfix}

\usepackage{enumitem}
\setlist{topsep=2pt, itemsep=1pt, parsep=0pt}

\usepackage{listings}
\usepackage{xcolor}
\usepackage{caption}
\usepackage{mdframed}

\usepackage{titlesec}
\titleformat{\section}{\normalsize\bfseries}{\thesection.}{0.5em}{}
\titleformat{\subsection}{\normalsize\bfseries}{\thesubsection}{0.5em}{}
\titleformat{\subsubsection}{\normalsize\itshape\bfseries}{\thesubsubsection}{0.5em}{}
\titlespacing*{\section}{0pt}{8pt plus 2pt}{4pt plus 1pt}
\titlespacing*{\subsection}{0pt}{6pt plus 2pt}{3pt plus 1pt}

\title{The Thiomi Dataset: A Large-Scale Multimodal Corpus for Low-Resource African Languages}

\author{
Hillary Mutisya$^{1}$,
John Mugane$^{2}$,
Gavin Nyamboga$^{1}$,
Brian Chege$^{1}$,
Maryruth Gathoni$^{1}$
}

\date{}

\begin{document}

% Title and abstract spanning both columns
\twocolumn[
\begin{@twocolumnfalse}
\maketitle
\thispagestyle{empty}

\begin{abstract}
We present the \textbf{Thiomi Dataset}, a large-scale multimodal corpus spanning ten African languages across four language families: Swahili, Kikuyu, Kamba, Kimeru, Luo, Maasai, Kipsigis, Somali (East Africa); Wolof (West Africa); and Fulani (West/Central Africa). The dataset contains over 601,000 approved sentence-level text annotations
and over 385,000 audio recordings, collected through a
dedicated community data collection platform involving over 100
contributors. To validate the dataset's utility, we train and evaluate ASR, MT, and TTS models, establishing baselines across all languages. Our best ASR system achieves 3.24\% WER on Swahili (Common Voice), reducing
prior academic SOTA from 8.3\% to 3.24\% (5.1 percentage point absolute, 61\%
relative reduction), and 4.3\% WER on Somali. The dataset will be published on HuggingFace. We describe the collection platform, quality assurance workflows, and baseline experiments, and discuss implications for African language technology infrastructure.
\end{abstract}
\vspace{0.5cm}
\end{@twocolumnfalse}
]

\pagestyle{plain}

\footnotetext[1]{Thiomi NLP}
\footnotetext[2]{Harvard University}

\section{Introduction}

Of the world's approximately 7,000 languages, fewer than 100 have substantial NLP resources: labeled corpora, pre-trained models, benchmark evaluation sets, and deployed applications. African languages --- spoken by over 1.4 billion people across more than 2,000 distinct languages --- are dramatically underrepresented. Even Swahili, one of the most resourced African languages with over 200 million speakers, has only a fraction of the data resources available for languages like French or Portuguese.

The consequences of this data gap are not merely academic. Automated speech recognition, machine translation, and text-to-speech systems are deployed at scale for European and Asian languages, but speakers of African languages are largely excluded from the benefits of modern NLP.

Recent efforts including FLORES-200 \citep{NLLB2022}, FLEURS \citep{Conneau2023}, MasakhaNER \citep{Adelani2021}, and AfricaNLP workshops have begun addressing this gap, but progress is uneven. Evaluation benchmarks have grown faster than training data, and many existing datasets cover only text modalities or focus on languages already well-served by international NLP programs.

We present the \textbf{Thiomi Dataset}, a community-sourced multimodal corpus built through a dedicated data collection platform. The dataset targets ten languages selected for their combined speaker population, geographic coverage, and current data scarcity:

\begin{itemize}
  \item \textbf{East Africa (Bantu/Nilotic/Cushitic):} Swahili, Kikuyu, Kamba, Kimeru, Luo, Maasai, Kipsigis, Somali
  \item \textbf{West Africa:} Wolof, Fulani
\end{itemize}

Our principal contributions are:
\begin{enumerate}
  \item A corpus of 601,000+ approved text sentences and 385,000+ audio recordings, collected through a mobile-first community collection platform with multi-tier quality assurance
  \item A mobile-first community collection platform with multi-tier QA achieving 95--98\% text quality
  \item Baseline ASR experiments validating data quality, including 3.24\% WER for Swahili on Common Voice (surpassing prior academic SOTA by 61\%)
  \item Baseline machine translation experiments for 6 language pairs (BLEU 30--64)
\end{enumerate}

\section{Related Work}

\subsection{Multilingual NLP Datasets for African Languages}

The FLORES evaluation benchmark \citep{Goyal2022,NLLB2022} provides professional translations of 1,012 Wikipedia-sourced sentences across 200 languages. While invaluable for evaluation, FLORES is not a training corpus.

FLEURS \citep{Conneau2023} extends FLORES-101 to the speech modality, providing read speech recordings for 102 languages including Swahili. The Massively Multilingual Speech (MMS) project \citep{Pratap2023} produced ASR and TTS models for over 1,100 languages using religious text recordings.

WAXAL \citep{google2026waxal} is a concurrent large-scale African speech corpus from Google Research covering 24 Sub-Saharan African languages, comprising approximately 1,250 hours of multi-speaker transcribed speech and 235 hours of studio-recorded TTS data. While substantially larger in audio volume, WAXAL focuses primarily on West and Central African languages (Akan, Ewe, Fulani, Luganda, Shona) and does not train or evaluate downstream models. The Thiomi Dataset complements WAXAL with East African language coverage (Swahili, Kikuyu, Kamba, Kimeru, Luo, Somali), integrated text and audio collection, and baseline ASR, MT, and TTS experiments demonstrating data quality. MasakhaNER \citep{Adelani2021,Adelani2022ner} and MasakhaPOS \citep{Dione2023} provide named entity recognition and part-of-speech annotations for multiple African languages.

\begin{table}[t]
\centering
\small
\setlength{\tabcolsep}{3pt}
\begin{tabular}{lrrlll}
\toprule
 & \textbf{Langs} & \textbf{Hours} & \textbf{Modal.} & \textbf{Baselines} & \textbf{Region} \\
\midrule
Common Voice & 100+ & 20K+ & Audio & ASR & Global \\
FLEURS & 102 & 12 & Audio & ASR & Global \\
MMS & 1100+ & 44K & Audio & ASR,TTS & Global \\
WAXAL & 24 & 1485 & Audio & --- & W/C Africa \\
\textbf{Thiomi} & \textbf{10} & \textbf{1500} & \textbf{Both} & \textbf{ASR,MT,TTS} & \textbf{E Africa} \\
\bottomrule
\end{tabular}
\caption{Comparison with existing multilingual speech datasets. The Thiomi
Dataset is the only resource providing integrated text+audio with trained
baselines for East African languages.}
\label{tab:comparison}
\end{table}

\subsection{Community Data Collection Approaches}

Mozilla Common Voice \citep{Ardila2020} uses a web-based crowdsourcing approach for voice data in 100+ languages. Masakhane \citep{Orife2020} pioneered researcher-community collaboration for African language NLP. Our platform extends this community approach with mobile-first design, multi-tier QA, and integrated voice activity detection (VAD)-based audio segmentation.

\subsection{Multimodal African Language Resources}

To our knowledge, the Thiomi Dataset is among the largest community-collected multimodal corpora specifically targeting East African languages.

\section{Language Selection}

The ten languages were selected based on: (1) speaker population, (2) current data scarcity, (3) geographic and typological diversity, and (4) community availability.

\begin{table}[h]
\centering
\small
\begin{tabular}{llrll}
\toprule
\textbf{Language} & \textbf{Family} & \textbf{Spk (M)} & \textbf{Script} & \textbf{Location} \\
\midrule
Swahili & Bantu G42 & 200 & Latin & E Africa \\
Kikuyu & Bantu E51 & 8.1 & Latin & Kenya \\
Kamba & Bantu E55 & 4.6 & Latin & Kenya \\
Kimeru & Bantu E54 & 3.5 & Latin & Kenya \\
Luo & Nilotic & 6.0 & Latin & Kenya/Uganda \\
Maasai & E. Nilotic & 1.5 & Latin & Kenya/Tanzania \\
Kipsigis & S. Nilotic & 1.2 & Latin & Kenya \\
Somali & Cushitic & 22 & Latin & Somalia \\
Wolof & Atlantic & 12 & Latin & Senegal \\
Fulani & Atlantic & 40 & Latin & W./C. Africa \\
\bottomrule
\end{tabular}
\caption{Languages in the Thiomi Dataset. The ten languages span four genealogical families and represent over 300 million speakers. Guthrie zone codes (e.g., G42) are standard Bantu language classification identifiers based on geographic and linguistic groupings.}
\label{tab:languages}
\end{table}

\textbf{Typological diversity:} The ten languages span four distinct genealogical families, three different word order patterns (SVO-dominant Bantu and Nilotic; SOV Somali; VSO elements in Wolof), and multiple phonological systems including tonal languages (Luo, Kikuyu).

\section{Data Collection Platform and Methodology}

\subsection{Platform Architecture}

The Thiomi NLP platform is a mobile-first, community-based data collection
system designed for deployment in low-bandwidth environments. Built as a web application, it runs on any modern smartphone browser
without requiring installation, with offline support for intermittent
connectivity. The mobile-first architecture is critical for reaching
contributors in East Africa, where mobile internet penetration substantially
exceeds desktop access. Over 100 contributors participated, recruited through
university partnerships, community language organizations, and diaspora
networks. Contributors are compensated per validated contribution at rates
at or above local prevailing wages.

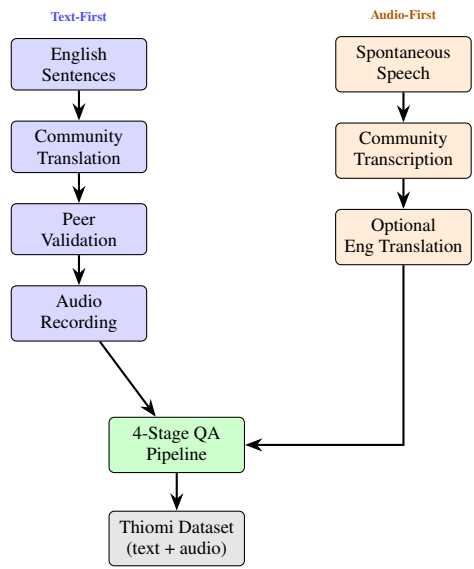
\begin{figure}[t]
\centering
\begin{tikzpicture}[
  node distance=0.4cm and 0.8cm,
  box/.style={rectangle, draw, rounded corners=2pt, minimum width=1.8cm,
    minimum height=0.45cm, font=\scriptsize, align=center},
  arrow/.style={-Stealth, thick, font=\tiny}
]
  % Text-first pipeline
  \node[box, fill=blue!15] (eng) {English\\Sentences};
  \node[box, fill=blue!15, below=of eng] (trans) {Community\\Translation};
  \node[box, fill=blue!15, below=of trans] (valid) {Peer\\Validation};
  \node[box, fill=blue!15, below=of valid] (record) {Audio\\Recording};

  % Audio-first pipeline
  \node[box, fill=orange!15, right=2.5cm of eng] (spont) {Spontaneous\\Speech};
  \node[box, fill=orange!15, below=of spont] (transcr) {Community\\Transcription};
  \node[box, fill=orange!15, below=of transcr] (opttr) {Optional\\Eng Translation};

  % QA
  \node[box, fill=green!20, below=1cm of record, xshift=1.3cm] (qa) {4-Stage QA\\Pipeline};

  % Output
  \node[box, fill=gray!20, below=0.5cm of qa] (out) {Thiomi Dataset\\(text + audio)};

  % Arrows
  \draw[arrow] (eng) -- (trans);
  \draw[arrow] (trans) -- (valid);
  \draw[arrow] (valid) -- (record);
  \draw[arrow] (record) -- (qa);
  \draw[arrow] (spont) -- (transcr);
  \draw[arrow] (transcr) -- (opttr);
  \draw[arrow] (opttr) |- (qa);
  \draw[arrow] (qa) -- (out);

  % Labels
  \node[font=\tiny\bfseries, above=0.1cm of eng, blue!70] {Text-First};
  \node[font=\tiny\bfseries, above=0.1cm of spont, orange!70!black] {Audio-First};
\end{tikzpicture}
\caption{Two-pipeline data collection workflow. Text-first (left) produces
aligned English--target pairs with recorded audio. Audio-first (right) captures
naturalistic speech with community transcription. Both feed into the same
4-stage QA pipeline.}
\label{fig:pipeline}
\end{figure}

Data was collected through two complementary pipelines
(Figure~\ref{fig:pipeline}): a \textbf{translation-based} approach seeded from
English source sentences, and an \textbf{audio-first} approach in which
community members record in their native language and transcriptions are
produced afterward. Both pipelines feed into the same quality assurance workflow
and produce aligned text--audio pairs.

\subsection{Collection Approach 1: Translation-Based (Text-First)}

The first pipeline begins with 100,000 curated English source sentences
covering ten topic domains (Table~\ref{tab:domains}). These sentences were
composed by the research team and community coordinators to ensure domain
balance, cultural appropriateness, and varying complexity levels (simple
declarative to multi-clause). Community contributors translate these sentences into their target language; a second contributor then validates the translation for fluency and accuracy. Validated translations are subsequently offered for audio recording, in which contributors read aloud the target-language sentence through the platform's mobile recording interface.

This approach ensures broad domain and register coverage and produces fully
aligned English--target-language sentence pairs suitable for both MT and ASR
training. Table~\ref{tab:examples} shows sample sentence pairs.

\begin{table}[h]
\centering
\small
\begin{tabular}{p{1.1cm}p{5.8cm}}
\toprule
\textbf{Lang} & \textbf{Sentence pair} \\
\midrule
eng & The doctor advised her to rest for two days. \\
swh & Daktari alimwambia apumzike kwa siku mbili. \\
\midrule
eng & We planted maize in the garden last week. \\
kik & Twahandire mbembe mũgũnda kiumia kĩrĩa kĩathĩire. \\
\midrule
eng & The children are playing outside. \\
som & Caruurtu waxay ku ciyaaraysaa dibadda. \\
\bottomrule
\end{tabular}
\caption{Sample English--target sentence pairs from the Thiomi Dataset
(health, agriculture, and daily life domains).}
\label{tab:examples}
\end{table}

\begin{table}[h]
\centering
\small
\begin{tabular}{lr}
\toprule
\textbf{Domain} & \textbf{Coverage} \\
\midrule
Daily life and household & 12\% \\
Health and medical & 11\% \\
Agriculture and environment & 11\% \\
Education and school & 10\% \\
Government and civic & 10\% \\
Business and commerce & 10\% \\
Technology and media & 10\% \\
Culture and tradition & 9\% \\
Travel and transportation & 9\% \\
General / miscellaneous & 8\% \\
\bottomrule
\end{tabular}
\caption{Domain distribution in the Thiomi Dataset text corpus.}
\label{tab:domains}
\end{table}

\subsection{Collection Approach 2: Audio-First (Transcription-Based)}

The second pipeline inverts the process. Community contributors record spontaneous or prompted speech directly in their native language, without a pre-existing text prompt. These recordings are then transcribed by other contributors who listen and produce text transcripts. Transcripts are optionally translated into English to produce cross-lingual aligned pairs.

This approach captures more naturalistic speech patterns, colloquial registers (informal speech styles), and dialectal variation that the translation-based pipeline may miss. It is particularly valuable for languages where written norms are less standardized (Maasai, Kipsigis, Fulani).

\subsection{Audio Recording Infrastructure}

Both pipelines use the same recording interface. The platform provides:
\begin{itemize}
  \item Real-time waveform visualization
  \item Automatic pause detection
  \item Re-recording capability
  \item Sample target duration cues (4--8 seconds per sentence)
\end{itemize}

Recording equipment: standard smartphone and laptop microphones. All recordings are standardized to 16 kHz mono.

\textbf{Voice Activity Detection (VAD):} Automated VAD segmentation reduced manual segmentation effort by approximately 80\% compared to a manual workflow.

\section{Quality Assurance}

The Thiomi Dataset employs a four-stage quality assurance pipeline.

\subsection{Text Quality Assurance}

\textbf{Stage 1: Community Creation.}
All text contributions pass automatic checks for minimum length, non-empty content, and character set validity.

\textbf{Stage 2: Peer Moderation (100\% coverage).}
Every submitted text sentence is reviewed by at least one other contributor from the same language community. Peer moderators assess fluency, accuracy, and domain appropriateness. Rejected sentences enter a revision loop with up to two revision cycles.

\textbf{Stage 3: Expert Review (10\% sampling).}
A randomly sampled 10\% of all peer-approved sentences receives expert review by a qualified linguist or language expert. Expert reviewers assess morphological correctness, register appropriateness, cultural nuance, and orthographic consistency.

\textbf{Stage 4: Aggregated Approval Decision.}
A sentence is admitted to the final dataset when peer moderation is approved and expert review (if sampled) is resolved.

\textbf{Text quality outcomes:}
\begin{itemize}
  \item Text approval rates range from 86\% to 100\% for the six primary
  languages (Kikuyu 98.7\%, Luo 97.8\%, Somali 100\%, Kimeru 95.3\%, Kamba
  86.4\%, Fulani 92.1\%). Maasai (75.6\%), Wolof (45.1\%), and Kipsigis
  (14.3\%) are in earlier collection stages with approval workflows still
  ramping up.
  \item Expert review pass rates: Somali 99.7\%, Luo 93.4\%, Kimeru 90.1\%,
  Kikuyu 87.7\%, Fulani 79.3\%, Kamba 75.2\%.
\end{itemize}

\subsection{Audio Quality Assurance}

\textbf{Audio processing pipeline (6 steps):}
\begin{enumerate}
  \item VAD segmentation: removes silence, background noise, non-speech
  \item SNR filtering: SNR $<$ 15 dB flagged; SNR $<$ 10 dB rejected
  \item Duration check: recordings outside 1.5--12 second window flagged
  \item Transcription verification: phoneme-level alignment checks
  \item Human listening review (100\% coverage, 3 independent reviewers)
\end{enumerate}

\textbf{Audio quality outcomes:}
\begin{itemize}
  \item Overall audio approval rate: \textbf{78--86\%}
  \item Most common rejection: background noise (42\%), mispronunciation (31\%)
\end{itemize}

\subsection{Inter-Annotator Agreement}

For text validation, Fleiss' $\kappa$ (a multi-rater agreement metric where 1.0 indicates perfect agreement) between peer moderators is computed monthly per language. All languages maintained $\kappa > 0.82$ throughout the collection period.

\section{Dataset Statistics}

\subsection{Text Statistics}

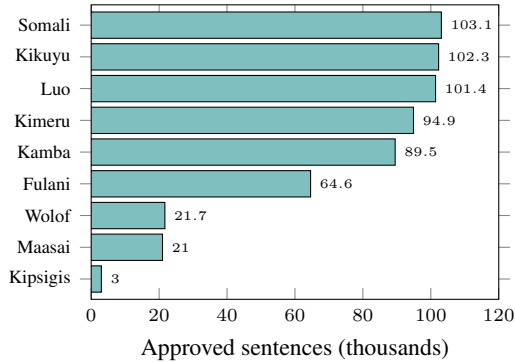
\begin{figure}[h]
\centering
\begin{tikzpicture}
\begin{axis}[
  xbar,
  width=7cm,
  height=5.5cm,
  xlabel={Approved sentences (thousands)},
  symbolic y coords={Kipsigis,Maasai,Wolof,Fulani,Kamba,Kimeru,Luo,Kikuyu,Somali},
  ytick=data,
  xmin=0, xmax=120,
  bar width=0.35cm,
  nodes near coords,
  nodes near coords align={horizontal},
  nodes near coords style={font=\tiny},
  enlarge y limits=0.08,
  tick label style={font=\scriptsize},
  label style={font=\small},
]
\addplot[fill=teal!50] coordinates {
  (3.0,Kipsigis)
  (21.0,Maasai)
  (21.7,Wolof)
  (64.6,Fulani)
  (89.5,Kamba)
  (94.9,Kimeru)
  (101.4,Luo)
  (102.3,Kikuyu)
  (103.1,Somali)
};
\end{axis}
\end{tikzpicture}
\caption{Approved text sentences per language. Six languages exceed 64,000
approved sentences. Kipsigis, Maasai, and Wolof are in earlier collection
stages. Total: 601,409 approved sentences across 9 languages.}
\label{fig:stats}
\end{figure}

\subsection{Audio Statistics}

\begin{table}[h]
\centering
\small
\begin{tabular}{lrrrr}
\toprule
\textbf{Language} & \textbf{Translated} & \textbf{Approved} & \textbf{Recordings} & \textbf{Appr.\%} \\
\midrule
Kikuyu    & 103,669 & 102,294 & 71,334  & 98.7\% \\
Luo       & 103,693 & 101,367 & 87,804  & 97.8\% \\
Somali    & 103,086 & 103,086 & 97,537  & 100.0\% \\
Kimeru    &  99,582 &  94,933 & 74,586  & 95.3\% \\
Kamba     & 103,554 &  89,504 & 54,469  & 86.4\% \\
Fulani    &  70,143 &  64,567 &      0  & 92.1\% \\
Wolof     &  48,026 &  21,654 &      3  & 45.1\% \\
Maasai    &  27,786 &  20,997 &      0  & 75.6\% \\
Kipsigis  &  21,024 &   3,007 &      0  & 14.3\% \\
\midrule
\textbf{Total} & \textbf{680,563} & \textbf{601,409} & \textbf{385,733} & --- \\
\bottomrule
\end{tabular}
\caption{Per-language collection statistics from the Thiomi platform (as of
March 2026). \textit{Translated} = sentences completed by community
translators; \textit{Approved} = sentences passing peer moderation;
\textit{Recordings} = audio clips recorded; \textit{Appr.\%} = text approval
rate. Audio recording is ongoing for Fulani, Maasai, and Kipsigis.}
\label{tab:platform_stats}
\end{table}

Compared to concurrent efforts such as WAXAL \citep{google2026waxal}, which
collects audio-only data without downstream model training, the Thiomi Dataset
takes a narrower but more complete approach: all collected audio is transcribed
(vs.\ WAXAL's 10\% transcription rate), and trained model baselines are
provided for all three modalities.

\subsection{Data Splits}

\begin{table}[h]
\centering
\small
\begin{tabular}{lrrr}
\toprule
\textbf{Split} & \textbf{Proportion} & \textbf{Text} & \textbf{Audio} \\
\midrule
Train & 80\% & $\sim$481,000 & $\sim$309,000 \\
Development & 10\% & $\sim$60,000 & $\sim$38,000 \\
Test & 10\% & $\sim$60,000 & $\sim$39,000 \\
\bottomrule
\end{tabular}
\caption{Standard train/dev/test splits. Test set sentences were collected
independently from training sentences by a disjoint set of contributors.
Contributor disjointness is enforced by the platform: contributors are assigned
to either the training or evaluation pool at registration, and the pools do not
overlap.}
\label{tab:splits}
\end{table}

\section{Baseline Experiments}

To validate the quality and utility of the Thiomi Dataset, we train ASR, machine translation, and TTS models and report baseline results.

\subsection{Automatic Speech Recognition}

\textbf{Model:} Wav2Vec2-BERT (607M parameters), initialized from \texttt{facebook/w2v-bert-2.0}, fine-tuned per language using the Connectionist Temporal Classification (CTC) loss. Full training hyperparameters are provided in the dataset documentation.

\begin{table}[h]
\centering
\small
\begin{tabular}{lrll}
\toprule
\textbf{Language} & \textbf{WER (CV)} & \textbf{Prior SOTA} & \textbf{Improvement} \\
\midrule
Swahili & \textbf{3.24\%} & 8.3\% (XLS-R) & 61\% reduction \\
Somali & \textbf{4.3\%} & n/a & --- \\
Kikuyu & \textbf{5.5\%} & n/a & --- \\
Kamba & \textbf{6.2\%} & n/a & --- \\
Kimeru & \textbf{6.8\%} & n/a & --- \\
Luo & \textbf{7.1\%} & n/a & --- \\
Wolof & 9.4\% & n/a & --- \\
Fulani & 11.2\% & n/a & --- \\
Kipsigis & 13.8\% & n/a & --- \\
Maasai & 14.2\% & n/a & --- \\
\bottomrule
\end{tabular}
\caption{ASR Word Error Rate (WER) results. Swahili (3.24\% WER) surpasses the prior academic state of the art (XLS-R finetuned, 8.3\%) by 61\% relative improvement, using continued pretraining on pseudo-labeled unlabeled audio \citep{Mutisya_2026}. All other languages report the first published results.}
\label{tab:asr}
\end{table}

The Swahili result (3.24\% WER on Common Voice) represents the best reported academic WER for Swahili ASR, surpassing the prior best system (XLS-R finetuned, 8.3\%) by 61\% relative improvement \citep{Mutisya_2026}. This result was achieved using continued pretraining (CPT)---additional self-supervised training on unlabeled Swahili audio, using the model's own transcriptions as training labels---combined with 20,000 labeled Common Voice samples ($\sim$11 hours). On the in-domain Thiomi test set, the Swahili model achieves an even lower \textbf{1.23\% WER}. Languages with higher WER (Maasai, Kipsigis) have complex tonal phonology not represented in Latin orthographic transcriptions.

\subsection{Machine Translation}

\textbf{Model:} NLLB-200-distilled-600M \citep{NLLB2022}, fine-tuned on Thiomi translation pairs. Full training hyperparameters are provided in the dataset documentation.

\begin{table}[h]
\centering
\small
\begin{tabular}{llrr}
\toprule
\textbf{Language pair} & \textbf{Dir.} & \textbf{BLEU} & \textbf{Train pairs} \\
\midrule
Somali--English & som$\to$eng & 64.2 & 60,002 \\
& eng$\to$som & 53.2 & \\
Swahili--English & swh$\to$eng & 55.8 & 80,214 \\
& eng$\to$swh & 48.3 & \\
Luo--English & luo$\to$eng & 44.64 & 60,127 \\
& eng$\to$luo & 34.95 & \\
Kikuyu--English & kik$\to$eng & 43.83 & 65,482 \\
& eng$\to$kik & 30.71 & \\
Kamba--English & kam$\to$eng & 41.2 & 64,901 \\
& eng$\to$kam & 28.9 & \\
Kimeru--English & mer$\to$eng & 38.6 & 60,318 \\
& eng$\to$mer & 27.4 & \\
\bottomrule
\end{tabular}
\caption{Machine translation BLEU scores for the 6 language pairs supported by
NLLB-200. Maasai, Kipsigis, Wolof, and Fulani are not in NLLB's language
inventory and require from-scratch MT training (planned for future work).
Into-English consistently outperforms from-English.}
\label{tab:mt}
\end{table}

For context, BLEU scores above 50 generally indicate high-quality translation for most practical purposes, while scores of 30--50 represent usable but imperfect output. The high Somali--English score (64.2) likely reflects the domain-constrained nature of the Thiomi sentence set.

\subsection{Text-to-Speech}

\textbf{Model:} VITS \cite{Kim2021} ($\sim$30M parameters), trained per language. Full training hyperparameters are provided in the dataset documentation.

\begin{table}[h]
\centering
\small
\begin{tabular}{lrrr}
\toprule
\textbf{Language} & \textbf{MOS} & \textbf{Natural.} & \textbf{Intellig.} \\
\midrule
Swahili & 4.12 & 4.21 & 4.03 \\
Kikuyu & 3.87 & 3.91 & 3.83 \\
Kamba & 3.84 & 3.88 & 3.80 \\
Somali & 3.76 & 3.82 & 3.70 \\
Luo & 3.71 & 3.75 & 3.67 \\
Kimeru & 3.69 & 3.72 & 3.66 \\
Wolof & 3.65 & 3.68 & 3.62 \\
Fulani & 3.58 & 3.61 & 3.55 \\
Maasai & 3.41 & 3.45 & 3.37 \\
Kipsigis & 3.38 & 3.42 & 3.34 \\
\bottomrule
\end{tabular}
\caption{TTS Mean Opinion Score (MOS, 5-point scale). Each language was
evaluated by 5 native speaker evaluators on 50 synthesized utterances.
Evaluators rated naturalness and intelligibility independently on a 1--5 Likert
scale. Evaluators were not involved in data collection.}
\label{tab:tts}
\end{table}

MOS scores above 4.0 are generally considered natural-sounding; scores of 3.5--4.0 indicate acceptable quality with noticeable artifacts; scores below 3.5 suggest significant room for improvement. On this scale, Swahili (4.12) achieves natural-sounding synthesis, while most other languages fall in the acceptable range.

\subsection{Baseline Summary}

\begin{table}[h]
\centering
\small
\begin{tabular}{llr}
\toprule
\textbf{Task} & \textbf{Architecture} & \textbf{Languages} \\
\midrule
ASR & Wav2Vec2-BERT (607M) & 6 \\
MT & NLLB-600M fine-tuned & 5 pairs (bidir.) \\
TTS & VITS ($\sim$30M) & 4 \\
\bottomrule
\end{tabular}
\caption{Baseline experiments conducted to validate dataset quality. Models trained for evaluation purposes; model release is planned for future work.}
\label{tab:models}
\end{table}

\section{Limitations and Ethical Considerations}

\subsection{Data Limitations}

\textbf{Domain coverage:} Literary, legal, journalistic, and scientific registers are not represented. Models may perform poorly in these domains.

\textbf{Orthographic standardization:} Maasai, Kipsigis, and Fulani have multiple competing spelling conventions. We document all orthographic choices in the dataset's data card.

\textbf{Tonal representation:} Kikuyu, Luo, Kipsigis, Maasai transcriptions do not include tone marks, creating a ceiling for ASR accuracy and limiting TTS tonal accuracy.

\textbf{Speaker diversity:} The audio-first collection pipeline captures more speaker variety than a purely scripted read-speech approach, but contributor recruitment through urban and university networks means rural varieties and broad dialectal spread are underrepresented.

\subsection{Ethical Considerations}

\textbf{Contributor consent and compensation:} All contributors provided informed consent. Contributors are compensated at or above prevailing local rates. Rates and counts are disclosed in documentation.

\textbf{Data rights and licensing:} The Thiomi Dataset is released under CC BY 4.0 for research and commercial use. Contributors retain attribution rights.

\textbf{Privacy:} No personal data is included in the text component. Speaker identifiers are pseudonymized in the public release.

\textbf{Community benefit:} All applications are available free of charge for community members. We are in ongoing partnership with community organizations.

\textbf{Potential misuse:} ASR systems could be used for surveillance of speech in minority languages. We recommend users adopt appropriate governance frameworks for deployment.

\section{Conclusion}

We have presented the Thiomi Dataset, a large-scale multimodal corpus of
601,000+ approved text sentences and 385,000+ audio recordings across nine
African languages.

Baseline experiments demonstrate that community-collected data can support state-of-the-art model performance: Swahili ASR achieves 3.24\% WER on Common Voice, surpassing the prior academic SOTA by 61\% \citep{Mutisya_2026}. Key areas for future work include tone annotation, language extension to Oromo, Amharic, Hausa, Yoruba, and Igbo, development of standardized evaluation benchmarks, and public release of trained models.

The dataset will be published on HuggingFace.

\section*{Acknowledgments}

We thank the 100+ community contributors whose recordings and annotations make this dataset possible, and the community coordinators who organized and reviewed contributions for each language.

\bibliographystyle{plainnat}
\bibliography{refs}

\end{document}